\documentclass{article}

\usepackage{arxiv}

\usepackage[utf8]{inputenc} 
\usepackage[T1]{fontenc}    
\usepackage{hyperref}       
\usepackage{url}            
\usepackage{booktabs}       
\usepackage{amsfonts}       
\usepackage{nicefrac}       
\usepackage{microtype}      

\usepackage{algorithm,algorithmic}
\newcommand*\rot{\rotatebox{90}}
\usepackage{caption}
\usepackage{graphicx}
\usepackage{amsmath,amssymb} 
\usepackage{color}
\usepackage{authblk}

\title{Object Segmentation using Pixel-wise Adversarial Loss}

\author[1,2,3]{Ricard Durall}
\author[1]{Franz-Josef Pfreundt}
\author[4]{Ullrich K\"{o}the}
\author[1,5]{Janis Keuper}

\affil[1]{Fraunhofer ITWM, Germany}
\affil[2]{IWR, University of Heidelberg, Germany}
\affil[3]{Fraunhofer Center Machine Learning, Germany}
\affil[4]{Visual Learning Lab Heidelberg, Germany}
\affil[5]{Institute for Machine Learning and Analytics, Offenburg University, Germany}

\renewcommand{\headeright}{}
\renewcommand{\undertitle}{}

\begin{document}

\maketitle

\begin{abstract}
		Recent deep learning based approaches have shown remarkable success on object segmentation tasks. 
		However, there is still room for further improvement. 
		Inspired by generative adversarial networks, we present a generic end-to-end adversarial approach, 
		which can be combined with a wide range of existing semantic segmentation networks to improve their segmentation performance. 
		The key element of our method is to replace the commonly used 
		binary adversarial loss with a high resolution pixel-wise loss.  
		In addition, we train our generator employing stochastic weight averaging fashion, 
		which further enhances the predicted output label maps leading to state-of-the-art results. We show, that 
		this combination of pixel-wise adversarial training and weight averaging leads to  
		significant and consistent gains in segmentation performance, compared to the baseline models.
\end{abstract}

\section{Introduction}
\label{sec:introduction}
The semantic segmentation task consists of assigning a pre-defined class label to each pixel in an image. 
Due to its multi-domain versatility, it has become an important application in in the field of computer vision. Most of current state-of-the-art methods\,
\cite{long2015fully,ronneberger2015u,badrinarayanan2015segnet,lin2017refinenet,zhao2017pyramid,peng2017large,chen2018deeplab} rely on convolutional neural networks (CNN) approaches as a key part of their implementation. 
Nevertheless, semantic segmentation remains challenging: The reduced amount of available labeled pixel-wise data (i.e., each pixel has been annotated), the lack of context information, the huge variety of scenarios in which segmentation can be applied, are just a few examples that make it still hard to solve the segmentation task.

In this paper, we address the task of object segmentation by proposing an adversarial learning scheme which adds a pixel-wise adversarial loss to the classical topology. Inspired by generative adversarial networks (GAN)\,\cite{goodfellow2014generative}, which are being employed in quite a number of different fields, we use a discriminative network to generate additional information useful for the segmentation task. In our GAN-based semantic segmentation method, we can distinguish between the generative network (generator) and the discriminative network (discriminator). On one side, the generator creates fine label maps by optimizing an objective function that combines a multi-class cross entropy loss together with the adversarial loss. On the other side, the discriminator tries to determine whether a given input segmentation map comes from the training data (real) or belongs to the generated data (fake).

\begin{figure}
\centering
  \includegraphics[width=0.7\linewidth]{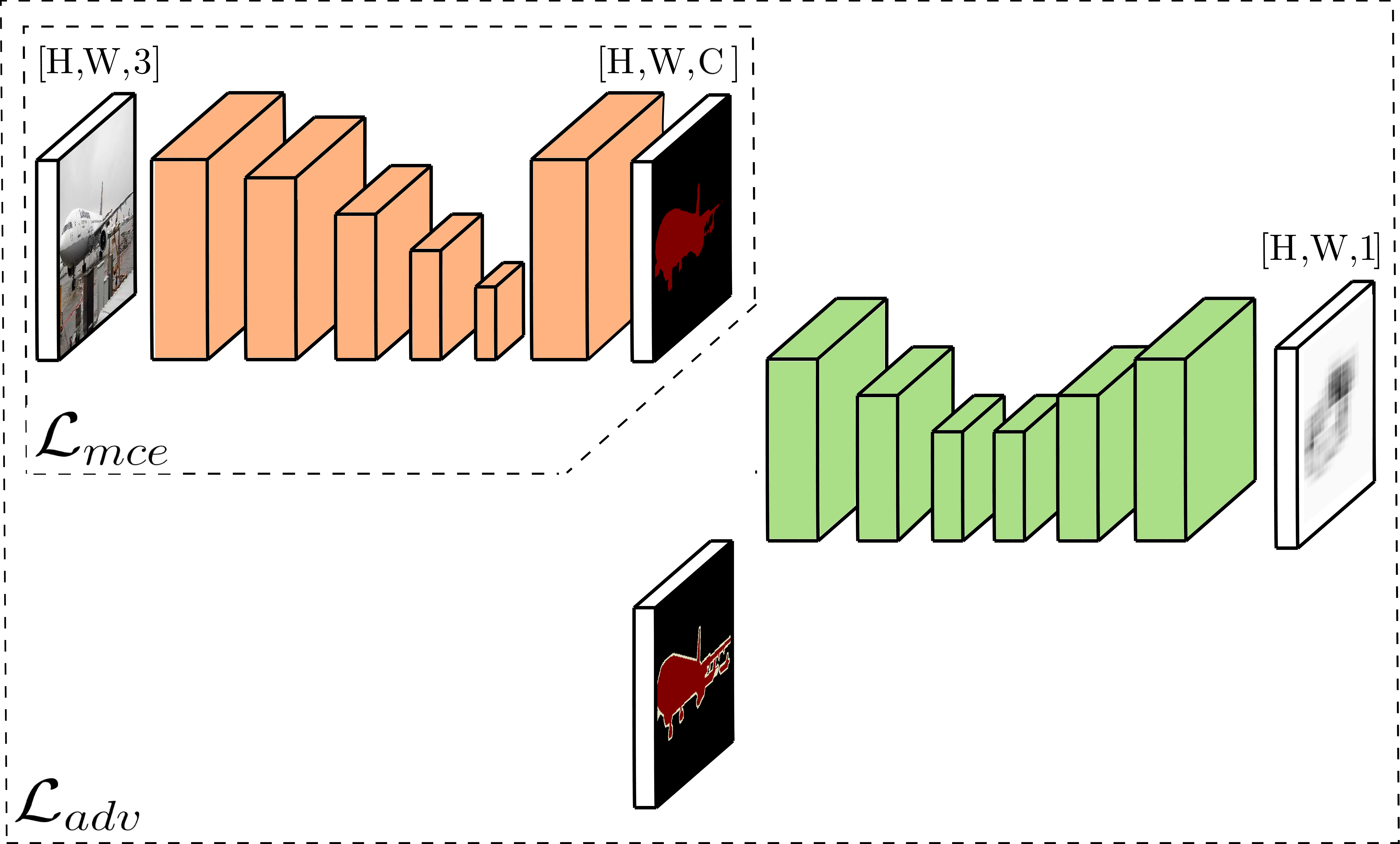}
  \caption{The segmentation network (orange/left) is fed with RGB images as input, and produces pixel-wise predicted maps as output. The discriminator (green/right) network is fed with the predicted maps or with the ground-truth label maps and determines if they are either real or false at pixel-wise level. }
  \label{fig:main}
\end{figure}

Summarizing, the contributions of our work are the following

\begin{itemize}
\item  We present a generic adversarial framework that leads to more accurate segmentation results than standalone segmentator.

\item  We introduce a simple encoder-decoder structure together with the pixel-wise loss function which define the discriminator.

\item The generator uses a weight averaging technique that allows traditional machine learning optimizers such as Stochastic Gradient Descent (SGD) to find broader minima with almost no extra computational cost.

\item Our experimental results show an improvement in labeling accuracy on both Pascal VOC 2012 and SegTrack-v2 datasets.
\end{itemize}

The rest of the paper is organized as follows. In the next section, we review related methods for semantic segmentation. In Section III, we introduce our approach, where we first provide a brief background of generative adversarial networks, and then we describe the design and methodology of our proposed architecture. Experimental results are shown in Section IV, where we report our results in Pascal VOC 2012 and SegTrack-v2 datasets. In Section V and VI is presented an experimental ablation study and the conclusions, respectively.

\section{Related Work}

\subsection{Generative Adversarial Networks}
Goodfellow et al. proposed an adversarial framework\,\cite{goodfellow2014generative} capable of learning deep generative models. It can be described as a minmax game between the generator $G$, which learns how to generate samples which resemble real data, and a discriminator $D$, which learns to discriminate between real and fake data. Throughout this process, $G$ indirectly learns how to model the input image distribution $p_{data}$ by taking samples $z$ from a fixed distribution $p_{z}$ (e.g. Gaussian)  and forcing the generated samples  $G(z)$ to match the natural images $x$. The objective loss function is defined as

\begin{align}
\begin{split}
	\min_{G} \max_{D} \mathcal{L}(D,G) =\,& \mathbb{E}_{\mathrm{\mathbf{x}} \sim p_{\mathrm{data}}} \left[ \log \left(D(x)\right) \right] \,+\, \mathbb{E}_{z \sim p_{z}}[\log(1-D(G(z))].
\end{split}
\end{align}

GANs have drawn significant attention from the computer vision community. Numerous works such as\,\cite{radford2015unsupervised,salimans2016improved,arjovsky2017wasserstein,gulrajani2017improved} have further extended and improved the original vanilla GAN\,\cite{goodfellow2014generative}. Moreover, it has been used in a wide variety of applications including image generation\,\cite{radford2015unsupervised,karras2017progressive}, domain adaptation\,\cite{isola2017image,zhu2017unpaired,hoffman2017cycada,tsai2018learning}, object detection\,\cite{wang2017fast,li2017perceptual}, video applications\,\cite{mathieu2015deep,vondrick2016generating,tulyakov2017mocogan} and semantic segmentation\,\cite{luc2016semantic,souly2017semi,xue2018segan,hung2018adversarial}.

\subsection{Semantic Segmentation} 

CNN-based approaches have become very popular within the computer vision field. Semantic segmentation has not been an exception and many promising ideas have been based on CNNs. Some of their  key contributions are the introduction of fully convolutional networks\,\cite{long2015fully,badrinarayanan2015segnet}, the usage of pre-trained models (mostly based on ImageNet\,\cite{deng2009imagenet}) and the implementation of long and short skip connections for improving the gradient flow\,\cite{ronneberger2015u}. As a result, much accurate semantic segmentation results have been achieved. More recent state-of-the-art results, have employed advanced typologies\,\cite{lin2017refinenet,zhao2017pyramid,peng2017large,chen2018deeplab}  along with new proposals such dilated convolution\,\cite{yu2015multi} to increase the receptive field size, improved pyramid techniques \cite{zhao2017pyramid} and even new loss functions like\,\cite{luc2016semantic,hung2018adversarial,matthew2018lovasz}.

\subsection{Adversarial training for Semantic Segmentation}

Unlike aforementioned methods regarding semantic segmentation, adversarial training for semantic segmentation uses an extra network, usually called discriminator, which provides feedback to the generator based on differences between the predictions and the ground-truth. Such a feedback is often modeled as an additional loss term. The same mechanism has been successfully applied to several semantic tasks such as medical image analysis and image-to-image translation\,\cite{isola2017image,zhu2017unpaired}.

In this work, we propose to use this adversarial methodology to learn the target distribution in a minmax fashion game between the generator and the discriminator. This approach is flexible enough to detect mismatches between model predictions and the ground-truth, without having to explicitly define these. As a result, more accurate segmentation masks than with traditional segmentation networks can be obtained.

\section{Method}
\subsection{Model Architecture}

As shown in Fig.\,\ref{fig:main}, the proposed model architecture consists of two parts: the segmentation network and the discrimination network.\\

\noindent \textbf{Segmentation/Generation network.} Given the generic attribute of the proposed framework, any network architecture designed for semantic segmentation task is suitable for being the segmentator. Its objective is to map an input image of dimension H$\times$W$\times$3, to probability maps of size H$\times$W$\times$C, being C the number of semantic classes.\\

\begin{figure}
  \centering
  \includegraphics[width=0.8\linewidth]{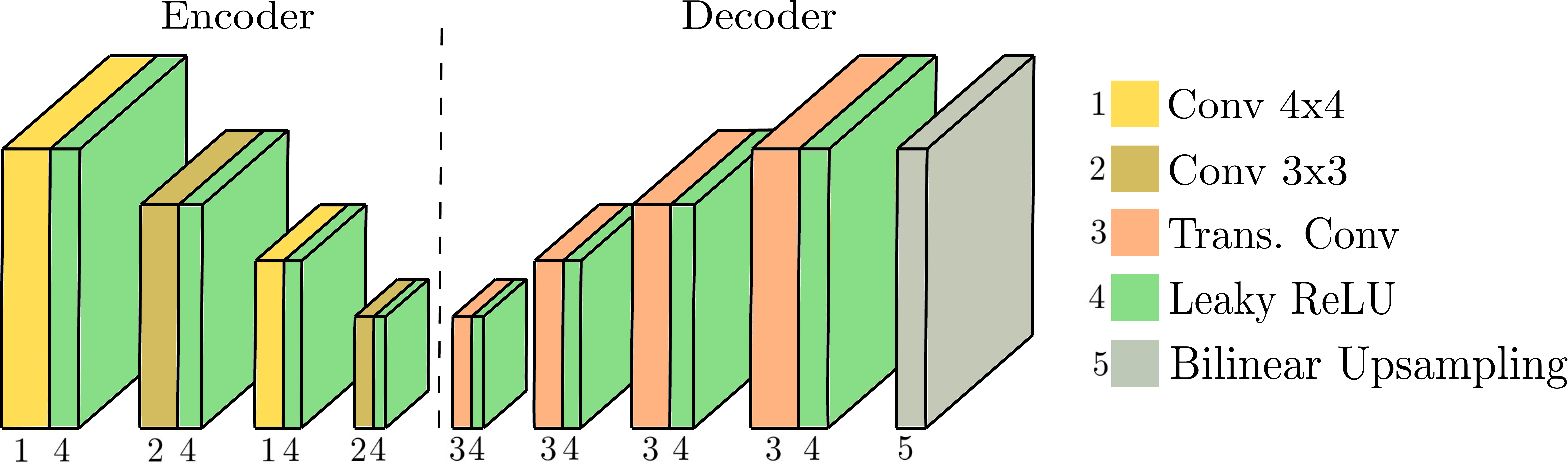}
  \caption{The schematics of the proposed discriminator network.}
  \label{fig:AE}
\end{figure}

\noindent \textbf{Discrimination network.} As discriminator, we introduce an encoder-decoder structure (see Fig.\,\ref{fig:AE}) which resembles an Auto-Encoder (AE) but without an explicit latent layer (bottleneck). 
It has H$\times$W$\times$C as an input image size and H$\times$W$\times$1 as an output confidence map size. 

\subsection{Adversarial Framework }

The generator and the discriminator are trained to optimize their corresponding loss function in an alternating fashion.\\

\noindent \textbf{Training Generator.} Given an input image $\mathrm{\mathbf{x}}$ and its label $y$,  the trainable parameters from the discriminator ($\theta_{\mathrm{disc}}$) are fixed and the generator parameters ($\theta_{\mathrm{gen}}$) are trained for one step. The objective function is defined as

\begin{align}
\begin{split}
	\mathcal{L}_{\mathrm{gen}}(\mathrm{\mathbf{x}},y;\theta_{\mathrm{gen}},\theta_{\mathrm{disc}}) = \,& \mathcal{L}_{\mathrm{mce}}(G(\mathrm{\mathbf{x}};\theta_{\mathrm{gen}}),y) \,+\, \lambda \, \mathcal{L}_{\mathrm{adv}}(G(\mathrm{\mathbf{x}};\theta_{\mathrm{gen}});\theta_{\mathrm{disc}}).
\end{split}
\end{align}

The loss function has two terms, the multi-class cross entropy ($\mathcal{L}_{\mathrm{mce}}$) and the adversarial term ($\mathcal{L}_{\mathrm{adv}}$). Furthermore, there is an extra term $\lambda$ that provides control on the influence that the adversarial part has in the generator. 
While the $\mathcal{L}_{\mathrm{mce}}$ uses the images $\mathrm{\mathbf{x}}$ as an input to encourage $G$ to predict the right class label at each pixel location, the $\mathcal{L}_{\mathrm{adv}}$ tries to fool the discriminator by producing feature maps $\hat{y}$ close to the ground-truth distribution labels $y$, in order to further to recover high frequency details which have been overlooked by the generator. The $\mathcal{L}_{\mathrm{mce}}$ can be seen as a modified version of binary cross entropy to tackle with multi-class problems. It is calculated by examining each pixel value individually and measures the distance (error) between each predicted pixel probability distribution and its real probability distribution over the classes. It can be defined as

\begin{align}
	\mathcal{L}_{\mathrm{mce}} = -\sum_{h,w} \sum_{c \in C} y^{(h,w,c)}log(\hat{y}^{(h,w,c)}).
\end{align}
The $\mathcal{L}_{\mathrm{adv}}$ can be seen as a binary cross entropy loss function applied to every single pixel. Such a multi-output discriminator was introduced by PatchGAN\,\cite{isola2017image}. The idea is that while training the generator, the outputs from the discrimintaor (pixel-wise level) have to be 1, therefore classified as true, and consequently having an error equal to 0. It can be defined as

\begin{align}
	\mathcal{L}_{\mathrm{adv}}(\hat{y};\theta_{\mathrm{disc}}) = \mathcal{L}_{\mathrm{bce}}(\hat{y},1;\theta_{\mathrm{disc}}) = -\sum_{h,w} \log (D(\hat{y};\theta_{\mathrm{disc}})^{(h,w)}).  
	\label{eq:1}
\end{align}

\noindent \textbf{Training Discriminator.} When the parameters from the generator ($\theta_{\mathrm{gen}}$) have already been updated, these are fixed and then we train on the $D$ for one step. This time the gradients come from discriminator's objective function defined as

\begin{align}
\begin{split}
	\mathcal{L}_{\mathrm{disc}}(\hat{y},y;\theta_{\mathrm{disc}}) =\,  \mathcal{L}_{\mathrm{bce}}(\hat{y},0;\theta_{\mathrm{disc}}) \,+ \mathcal{L}_{\mathrm{bce}}(y,1;\theta_{\mathrm{disc}}).
\end{split}
\label{eq:2}
\end{align}

In this part of the training, given $y$ the outputs from the discriminator have to be 1, therefore classified as true, and given $\hat{y}$ have to be 0, therefore classified as fake. Thus, on the one hand, the generator tries to minimize the adversarial loss ($\mathcal{L}_{\mathrm{bce}}(\hat{y},1;\theta_{\mathrm{disc}}) \rightarrow 0$, see Eq.\,\eqref{eq:1}). On the other hand, the discriminator aims to maximize it 
($\mathcal{L}_{\mathrm{bce}}(\hat{y},0;\theta_{\mathrm{disc}}) \rightarrow\, 0 \equiv \mathcal{L}_{\mathrm{bce}}(\hat{y},1;\theta_{\mathrm{disc}}) \rightarrow 1$, see Eq.\,\eqref{eq:2}).This adversarial training described, follows the minmax game written as

\begin{align}
	\min_{\theta_{\mathrm{gen}}} \max_{\theta_{\mathrm{disc}}} \mathcal{L}(\theta_{\mathrm{gen}},\theta_{\mathrm{disc}}).
\end{align}
Step by step, both the generator and the discriminator networks improve at their tasks. As a result, during inference the generative network produces label maps that are similar to the ground-truth.

\subsection{Pixel-wise Adversarial Loss}

\begin{figure}[h!]
\centering
  \includegraphics[width=0.7\linewidth]{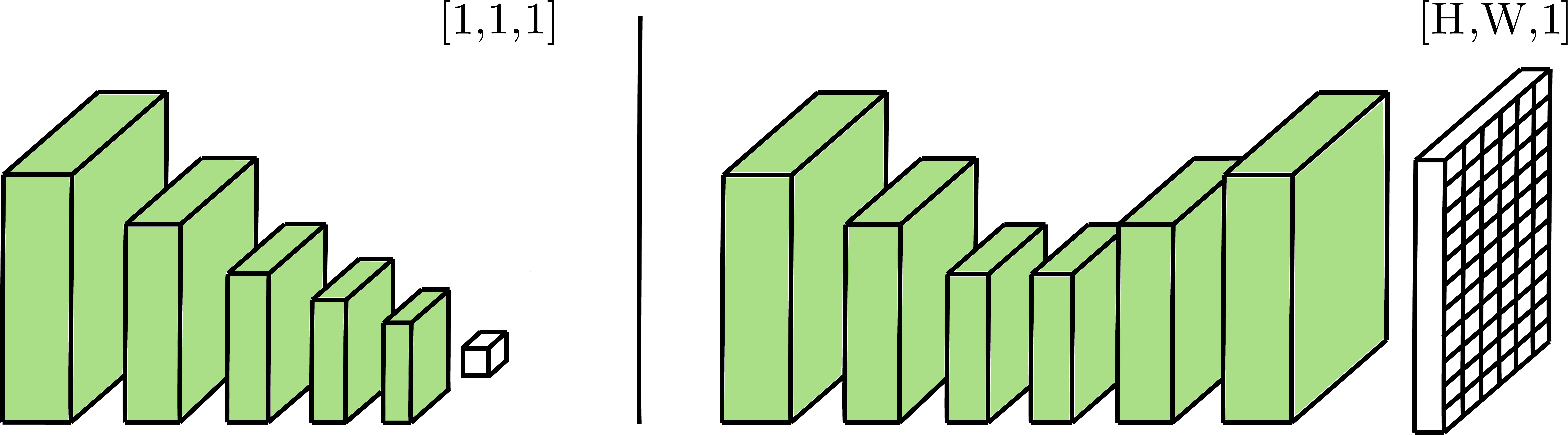}
  \caption{(Left) Vanilla adversarial loss structure. (Right) Pixel-wise adversarial loss.}
  \label{fig:loss}
\end{figure}

One key aspect in any adversarial setting is the composition of the adversarial loss. Previous publications use a loss based on traditional vanilla GAN, using a single binary decision per image. However, in this work we employ a different loss function, the pixel-wise adversarial loss. It makes a binary decision at pixel level. In other words, there are as many binary decisions as pixels in the input image (See in Fig.\,\ref{fig:loss}). Such loss function will provide more details and thus a more informative gradient can be back propagated.

\subsection{Averaging Weights}
Stochastic Weight Averaging (SWA)\,\cite{izmailov2018averaging} is a simple but effective algorithm for training any kind of neural network independently of its topology or loss function. The main idea consists of averaging of multiple points proposed by an SGD optimizer. In this way, it is possible to encounter a broader optima, which produces better generalization than conventional training methods.

Implementing SWA is rather straightforward and and not very compute intensive, we only need to keep updating the stochastic weight average $\theta_{\mathrm{swa}}$, written as
\begin{align}
	\theta_{\mathrm{swa}} \leftarrow \dfrac{\theta_{\mathrm{swa}} \cdot n+ \theta_{\mathrm{gen}}}{n + 1},
\end{align}

being $n$ the number of skipped iterations of the SWA per generator iteration and $\theta_{\mathrm{gen}}$ the generator parameters.

\section{Experiments}
In this section we present for a series of experiments evaluating the results on Pascal VOC 2012 and SegTrack-v2 datasets. We first give a detailed introduction of the experimental set up. Then, we discuss independently the results on the aforementioned dataset. 
\subsection{Experimental Settings}
In the experiments, we use a model pre-trained on the ImageNet dataset\,\cite{deng2009imagenet} and Microsoft COCO\,\cite{lin2014microsoft} as our segmentation baseline network. After initializing our model with these pre-trained parameters, we train on Pascal VOC 2012\,\cite{pascal-voc-2012} and SegTrack-v2\,\cite{li2013video} datasets. Our main evaluation metric is the mean Intersection-over-Union (mIoU) also known as Jaccard index and it is computed as

\begin{align}
	\mathrm{mIoU} =  \dfrac{1}{n}\sum_{n}\dfrac{ \vert  A_1 \bigcap A_2 \vert }{\vert  A_1 \bigcup A_2 \vert }, 
\end{align}
being $n$ the number of input samples, and $A_{1}$ and $A_{2}$ the ground-truth and predicted label maps respectively. All the experiments have been implemented in a single NVIDIA GeForce GTX 1080 GPU.

We conducted the experiments adopting a segmentation network based on\,\cite{hung2018adversarial}. It integrates a modified version of DeepLabv2\,\cite{chen2018deeplab} framework with ResNet-101\,\cite{he2016deep}, where no multi-scale layer is used and the atrous spatial pyramid pooling (ASPP) substitutes the last classification layer. Moreover, the strides from the last two convolutional layers are modified, so that, the output feature maps size are exactly $1/8$ of the input image size. Finally, dilated convolution in conv4 and conv5 layers are applied. Regarding the implementation of SWA, we maintain $\theta_{\mathrm{swa}}$ (only used in the generator)  and $\theta_{\mathrm{gen}}$ separately, and we use the $\theta_{\mathrm{swa}}$ at testing time. Algorithm\,\ref{alg:main} summarizes the training of the proposal model.

 \begin{algorithm}
 \caption{Training of the proposed architecture. All PASCAL VOC 2012 experiments in the paper used the default value $n_{\mathrm{iter}} = 20000$, $n=100$, $\alpha_{\mathrm{gen}} = \alpha_{\mathrm{disc}} = 0.00025$, $m = 16$.$^{*}$}
 \label{alg:main}
 \begin{algorithmic}[1]
  \STATE Require: $n_{\mathrm{iter}}$, number of iterations. $n$, number of skipped iterations of the SWA per generator iteration. $\alpha$'s, learning rate. $m$, batch size. $\hat{y}$, output from the generator.

  \STATE Require: $\theta_{\mathrm{gen}}^0$ , initial generator parameters. $\theta_{\mathrm{disc}}^0$, initial discriminator parameters. $\theta_{\mathrm{swa}}^0 =  \theta_{\mathrm{disc}}^0$, initial SWA parameters.
  \FOR {$i < n_{\mathrm{iter}}$}
  
  \STATE Sample $\{\mathrm{\mathbf{x}}^{(j)}\}_{j=0}^m$ a batch from images
  \STATE Sample $\{y^{(j)}\}_{j=0}^m$ a batch from masks
  
  \STATE \# Train generator $G$
  \STATE $\theta_{\mathrm{gen}} \leftarrow \theta_{\mathrm{gen}} + \alpha_{\mathrm{gen}} \nabla \mathcal{L}_{\mathrm{gen}}(\mathrm{\mathbf{x}},y;\theta_{\mathrm{gen}},\theta_{\mathrm{disc}})$
  \IF {$mod(i,n) = 0$}
  \STATE $\theta_{\mathrm{swa}} \leftarrow \dfrac{\theta_{\mathrm{swa}} \cdot n+ \theta_{\mathrm{gen}}}{n + 1}$
  \ENDIF
  
  \STATE \# Train discriminator $D$
  \STATE $\theta_{\mathrm{disc}} \leftarrow \theta_{\mathrm{disc}} + \alpha_{\mathrm{disc}} \nabla \mathcal{L}_{\mathrm{disc}}(\hat{y},y;\theta_{\mathrm{disc}})$

  \ENDFOR
 \end{algorithmic}
 \end{algorithm}
\vspace{-4mm}
$^{*}$ In Section SegTrack-v2 Dataset are further hyper-parameters introduced.

\subsection{Training}
Since our model is divided into two distinguishable parts, also two independent optimizers with their hyper-parameters are used during training.\\

\noindent \textbf{Generator}. The generator network integrates the SWA as optimization method. The learning rate used in the implementation has a polynomial decay in accordance with the number of iterations $k$, $\alpha_{\mathrm{gen}}^{k} =  \alpha_{\mathrm{gen}}^{0} \, (1-\dfrac{k}{\mathrm{max\_iter}})^{p}$, where the power decay $p$ = 0.9 and the initial learning rate $\alpha_{\mathrm{gen}}^{0} = 0.00025$. The momentum is set to 0.9 and the weight decay is $0.00005$.\\

\begin{figure}[h!]
\centering
  \includegraphics[width=0.9\linewidth]{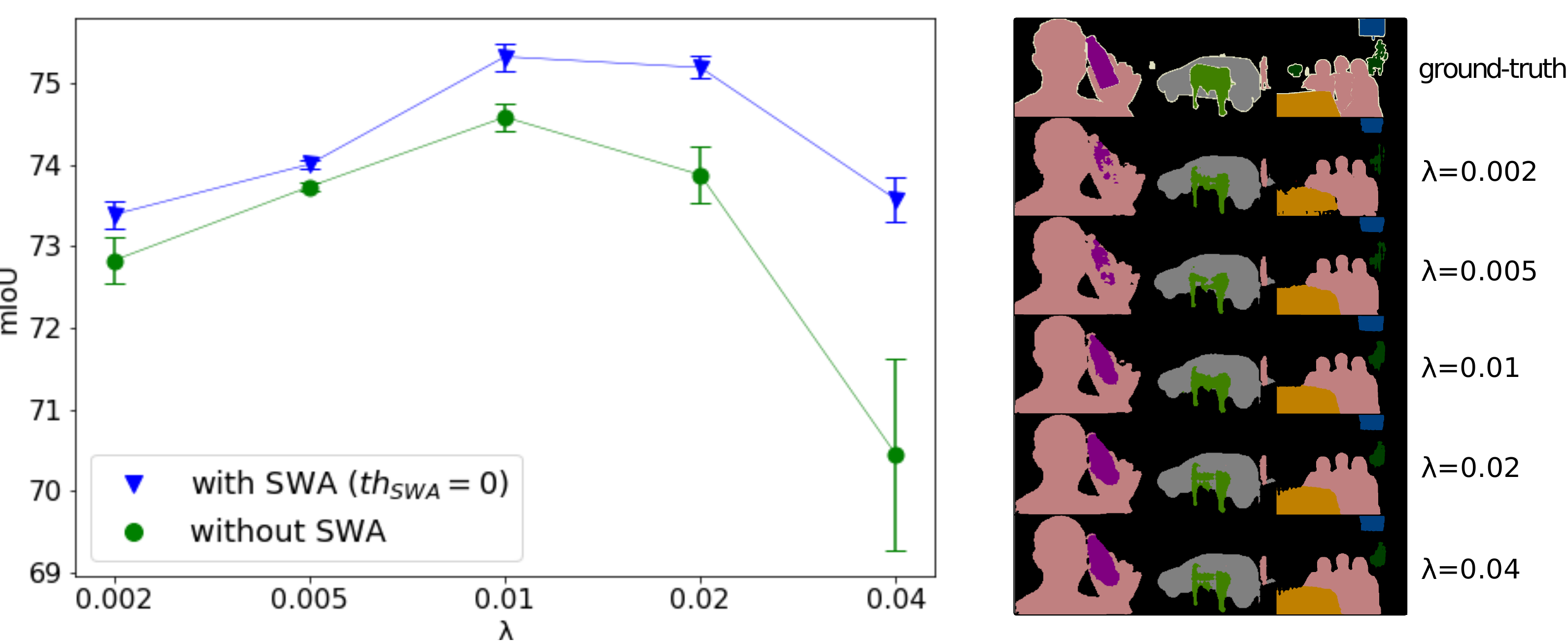}
  \caption{The evolution of mIoU score behaves in a similar manner for both systems (with and without SWA). Having a peak performance when $\lambda=0.01$. Results on Pascal VOC 2012 validation set.}
  \label{fig:lambda}
\end{figure}

\noindent  \textbf{Discriminator}. The discriminator uses Adam optimizer with learning rate starting at $10^{-4}$, $\beta_{1} = 0.9$, $\beta_{2} = 0.99$ and same annealing strategy as in the segmentation network. The adversarial loss $\mathcal{L}_{\mathrm{adv}}$ is weighted by the hyper-parameter $\lambda = 0.01$, which gives its best result (see Fig.\,\ref{fig:lambda}).

\subsection*{Hyper-Parameter Analysis}

In order to determine the optimal hyper-parameters, we have conducted an independent grid search for each one of them. On one hand, we have investigated the effects of $\lambda$. It is responsible to determine how much $\mathcal{L}_{\mathrm{adv}}$ will contribute to $\mathcal{L}_{\mathrm{gen}}$. However, it does not exist a unique solution. We have found that $\lambda$ is quite critical, since small modifications lead to big changes in mIoU score. 

On the other hand, $n$ and $th_{\mathrm{swa}}$ hyper-parameters control the effect of SWA. First we look for the optimal threshold percentage of iterations that should be used for the weight averaging $th_{\mathrm{swa}}$. We achieved best results when the weight averaging is applied from the beginning. Having a maximum score up to $73.1\%$, which is a gain of $1.0\%$  with respect to the baseline. (Note that for this test the discriminator is not involved). Second, after determining the optimum $th_{\mathrm{swa}}$ value, we have focused on the parameter $n$. This parameter has been found to have less fluctuations than $th_{\mathrm{swa}}$, resulting in a maximum relative variation between different $n$'s of $0.3\%$.

\begin{table*}[t!]
 \centering
 \caption{Per-class validation test mIoU in Pascal VOC 2012 dataset.}
 \resizebox{\textwidth}{!}{\begin{tabular}{ccccccccccccccccccccccc}
 		\hline
        Method & \rot{background} & \rot{aeroplane} & \rot{bicycle} & \rot{bird} & \rot{boat} 
        & \rot{bottle} & \rot{bus} & \rot{car} & \rot{cat} & \rot{chair} & \rot{cow} & \rot{dining table} 
        & \rot{dog} & \rot{horse} & \rot{motorbike} & \rot{person} & \rot{potted plant} & \rot{sheep} 
        & \rot{sofa} & \rot{train} & \rot{tv monitor} & mIoU\\
        \hline
        baseline  & 0.93 & 0.87 & 0.40 & 0.85 & 0.60 & 0.77 & 0.89 & 0.84 & 0.88 & 0.34 & 0.82 & 0.48 & 0.81 & 0.78 & 0.78 & 0.83 & 0.56 & 0.80 & 0.42 & \textbf{0.83} & 0.68 & 0.721 \\
        proposed  & \textbf{0.94} & \textbf{0.88} & \textbf{0.40} & \textbf{0.88} & \textbf{0.69} & \textbf{0.78} & \textbf{0.93} & \textbf{0.86} & \textbf{0.89} & \textbf{0.37} & \textbf{0.82} & \textbf{0.60} & \textbf{0.83} & \textbf{0.79} & \textbf{0.83} & \textbf{0.85} & \textbf{0.63} & \textbf{0.82} & \textbf{0.49} & 0.82 & \textbf{0.74} & \textbf{0.754} \\
        \hline
    \end{tabular}}    
    \label{table:miou_cl}
\end{table*}

\subsection*{Pascal VOC 2012 Dataset}

\begin{figure}[b!]
\centering
	\includegraphics[width=\linewidth]{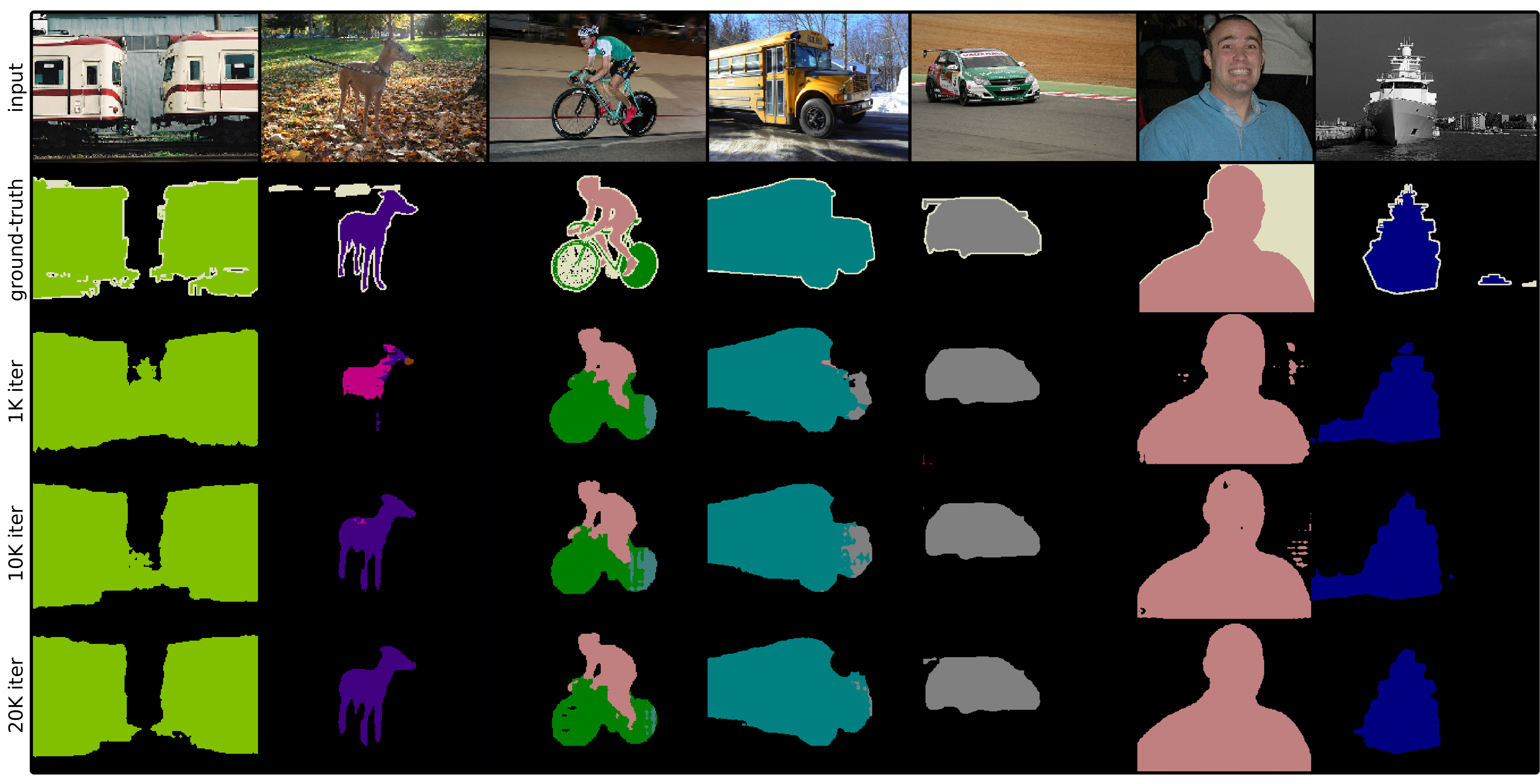}
  	\caption{Predicted masks at several stages of training in Pascal VOC 2012 dataset.}
  	\label{fig:training}
\end{figure}

We evaluate our approach on the evaluation set from Pascal VOC 2012 dataset\,\cite{pascal-voc-2012}. It is an extended benchmark for image segmentation that contains 1,464 and 1,449 images annotated with object instance contours for training and validation. These images have been slightly modified, by adding borders around the objects. Empirically, this procedure has been found beneficial for this dataset. With the further contribution of Segmentation Boundaries Dataset\,\cite{BharathICCV2011}, we end up with 10,528 and 1,449 images for training and validation. The data augmentation is used; random scaling and cropping operations are applied while training. We train our model for 20K iterations with batch size 10 (see Fig.\,\ref{fig:training}).

Our proposed framework clearly improves the baseline model, surpassing the mIoU among most of the classes and the overall. As can be seen in Table\,\ref{table:miou_cl}, when our proposed adversarial framework is fairly consistent, since it provides a steady boost in almost all the classes from the dataset. Table\,\ref{table:miouVoc} (right) reports in more detail the best result on the validation set and the improvement in the most common segmentation metrics respect to the baseline model.

We compare the proposed framework against other methods that also use adversarial training. Table\,\ref{table:comp} (left) shows how our approach has both a significant higher absolute gain and a higher mIoU. Note that both\,\cite{luc2016semantic,hung2018adversarial} have the same baseline model as we have. However, we achieve a final score $3.4\%$ higher than\,\cite{luc2016semantic} and  $0.5\%$  than\,\cite{hung2018adversarial}. The model suggested in\,\cite{hwang2018adversarial} uses a slightly different baseline version, since they do not remove the multi-scale fusion. As a result, their initial baseline is theoretically more advaced, therefore, stronger. Nevertheless, our method achieves a significant higher gain ($2.8\%$) compared to\,\cite{hwang2018adversarial}.

\begin{table}[t!]
\caption{Validation results in Pascal VOC 2012 dataset. (Left) Proposed model compared to other adversarial approaches. (Right) Common segmentation metrics. }
	\begin{minipage}{0.47\linewidth}
			\centering
 		\begin{tabular}{cccc}
 			\hline
     		&  Baseline & Adv. & Gain\\
   			\hline
    		\cite{luc2016semantic} & 71.8 & 72.0 & 0.2\\
    		\cite{hung2018adversarial} & 73.6 & 74.9 & 1.3\\
			\;\,\cite{hwang2018adversarial}\footnotetext{$^{*}$Complete DeepLabv2 scheme used as a generative network.}$^{*}$& 77.5 & 78.0 & 0.5\\
    		proposed & 72.1 & 75.4 & \textbf{3.3}\\
    		\hline
  		\end{tabular}
  		\label{table:comp}
 		
	\end{minipage}\hfill
	\begin{minipage}{0.5\linewidth}
		\centering
  		\begin{tabular}{cccccc}
  		\hline
    	& Overall Acc. & mAcc. & fwIoU & mIoU \\
    	\hline
   		baseline & 0.936 & 0.816 & 0.885 & 0.721\\
    	proposed  & 0.944 & 0.841 & 0.900 & \textbf{0.754}\\
    	\hline
  		\end{tabular}
  		\label{table:miouVoc}
	\end{minipage}
\end{table}

\subsection*{SegTrack-v2 Dataset}

We repeat the experiment, but this time we evaluate in SegTrack-v2\,\cite{li2013video}. It is a video segmentation dataset with full pixel-level annotations on multiple objects at each frame within each video. It contains 14 video sequences with 24 objects and 947 frames. Every frame is annotated with a pixel-level object mask. As instance-level annotations are provided for sequences with multiple objects, each specific instance segmentation is treated as separate problem. In our implementation, we have 16 classes, since we were treating repeated classes as one. There are no extra images but the same sort of augmentation applied in Pascal VOC 2012 is present in that experiment during training. We train our model for 5K iterations with a batch size 10.

\begin{table}[t!]
	\begin{minipage}{0.53\linewidth}
		\centering
  		\caption{Validation results in SegTrack-v2 dataset.}
  		\begin{tabular}{cccccc}
  		\hline
    	& Overall Acc. & mAcc. & fwIoU & mIoU \\
    	\hline
   		baseline & 0.989 & 0.871 & 0.980 & 0.812\\
    	proposed  & 0.991 & 0.872 & 0.983 & \textbf{0.831}\\
    	\hline
  		\end{tabular}
  		\label{table:conc2}
	\end{minipage}
	\begin{minipage}{0.4\linewidth}
  		\includegraphics[width=\linewidth]{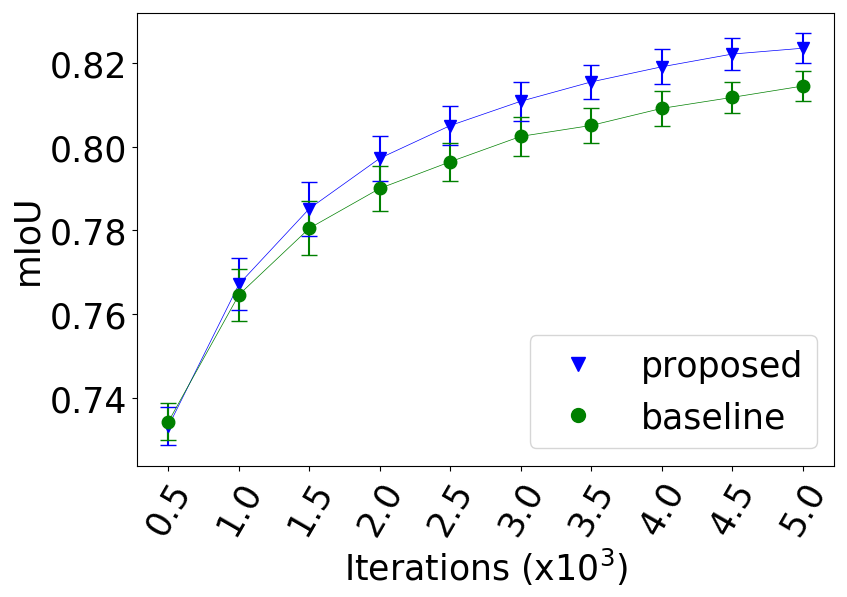}
  		\captionof{figure}{Evolution of mIoU from the baseline and the adversarial proposal during training in SegTrack-v2 dataset.}
  		\label{fig:segtrack}
	\end{minipage}\hfill
\end{table}

\begin{figure}
\centering
  	\includegraphics[width=0.9\linewidth]{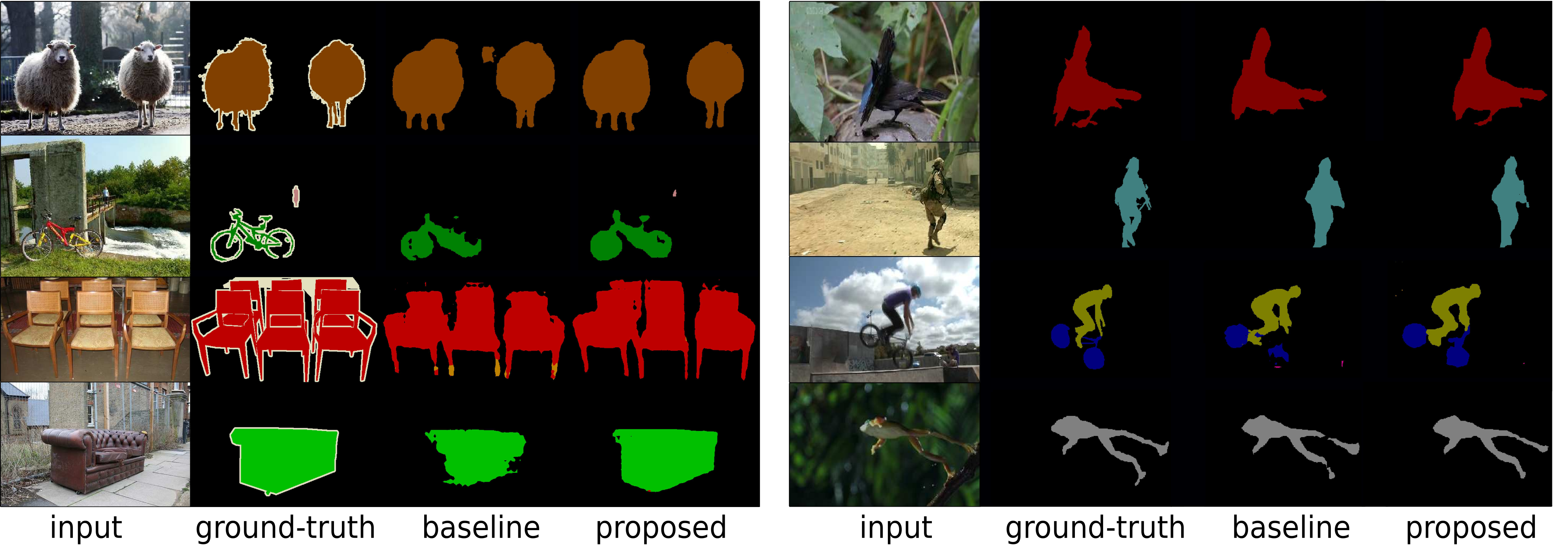}
  	\caption{Comparison of the predicted masks produced by the baseline and the proposed variant. (Left) Validation results in Pascal VOC 2012 dataset. (Right) Validation results in SegTrack-v2 dataset.}
  	\label{fig:comp}
\end{figure}

Fig.\,\ref{fig:segtrack} depicts the mIoU evolution during training. As it is expected, for both models the results get better over the training. However, using the adversarial set-up produces a boost on our model with respect to the baseline. Table\,\ref{table:conc2} provides an overview of the best results observed on the validation set using the proposal adversarial framework. It achieves a gain of $2.3\%$ with respect to the baseline. Some visual results are shown in Fig.\,\ref{fig:comp}, where top two rows belong to  the Pascal VOC 2012 and bottom two rows to SegTrack-v2 validation set.

\section{Ablation Study}
In this last section, we present further experiments that support the proposed adversarial framework. First, we conduct a comparison study between pixel-wise and standard loss. And then, we evaluate different segmentation networks together with our framework.

\subsection{Pixel-wise vs Standard Loss} 
The usage of adversarial techniques to further improve the segmentation masks is still an open discussion. As mentioned above, standard loss\,\cite{goodfellow2014generative} is by default the most widely extended. However, it is not always a reliable option. Empirically, we have observed the beneficial effects of pixel-wise loss (see Table\,\ref{table:conc}). It offers more steady, general (independent of datasets) and accurate results. This is possible thanks to the capacity of producing better gradients which flow from the adversarial loss to the segmentation network.

\begin{table}[h!]
	\begin{minipage}{0.48\linewidth}
		\centering
  		\caption{Comparison of the different adversarial loss functions.}
  		\begin{tabular}{ccc}
  		\hline
    	& Pascal VOC & SegTrackv2 \\
    	\hline
    	baseline & 0.721 & 0.812 \\
    	standard  & 0.742 & 0.547 \\
    	pixel-wise & \textbf{0.754} & \textbf{0.831} \\
    	\hline
  		\end{tabular}
  		\label{table:conc}
	\end{minipage}
	\begin{minipage}{0.4\linewidth}
  		\includegraphics[width=\linewidth]{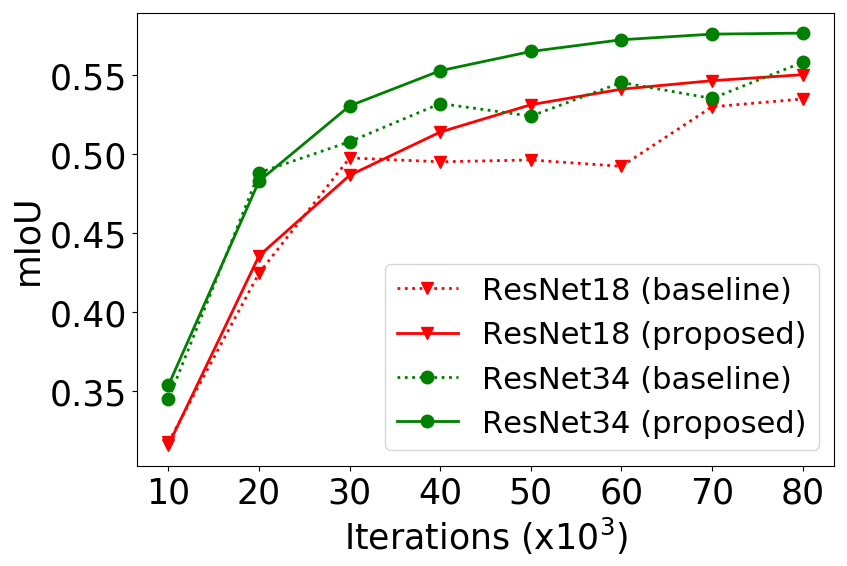}
  		\captionof{figure}{Comparison of different segmentation networks architecture.}
  		\label{fig:abal}
	\end{minipage}\hfill
\end{table}

\subsection{Flexibility capacity of our approach}
Finally, we have tested the versatility of our framework by using different architectures as a segmentation network. We replace our original segmentator topology with the DeepLabv3\,\cite{chen2017rethinking}. In this newer version, the ASPP module has been augmented with image-level feature to capture longer range information. Furthermore, it includes batch normalization parameters to facilitate the training.

We have run the adversarial extension within DeepLabv3. Such architecture has been evaluated in ResNet18 and ResNet34, without pre-trained model and making usage of the improved ASPP. Fig.\,\ref{fig:abal} depicts the results.  

\section{Conclusion}
In this paper, we present a novel end-to-end segmentation framework, making usage of a state-of-the-art pixel-wise adversarial loss for the discrimination network and the stochastic weight averaging. By training the discriminator in such an alternating fashion, all the experimental results get noticeably boosted, enhancing the final mIoU. Similar behaviour it is found on the ablation study, where we prove that our proposed adversarial scheme is effective and capable to lead to superior performances regardless of the segmentator.

\bibliographystyle{unsrt}
\bibliography{template}

\end{document}